\documentclass{article}
\usepackage{ijcai19}

\usepackage{times}
\usepackage{soul}
\usepackage{url}
\usepackage[hidelinks]{hyperref}
\usepackage[utf8]{inputenc}
\usepackage[small]{caption}
\usepackage{graphicx}
\usepackage{subfigure}
\usepackage{amsmath}
\usepackage{booktabs}
\usepackage{listings}
\usepackage{color}
\title{Adversarial Defense by Suppressing High-frequency Components}

\author{
Zhendong Zhang\and
Cheolkon Jung\And
Xiaolong Liang\\
\affiliations
Xidian Media Lab, Xidian University, China\\
\emails
zhd.zhang.ai@gmail.com,
zhengzk@xidian.edu.cn,
xlliang@stu.xidian.edu.cn
}

\begin{document}

\maketitle

\begin{abstract}
Recent works show that deep neural networks trained on image classification dataset bias towards textures. Those models are easily fooled by applying small high-frequency perturbations to clean images. In this paper, we learn robust image classification models by removing high-frequency components. Specifically, we develop a differentiable high-frequency suppression module based on discrete Fourier transform (DFT). Combining with adversarial training, we won the 5th place in the IJCAI-2019 Alibaba Adversarial AI Challenge. Our code is available online.
\end{abstract}

\section{Introduction}
Deep neural networks (DNNs) have achieved state-of-the-art performances on many tasks, such as image classifications. However, DNNs have been shown to be vulnerable to adversarial attacks \cite{szegedy2014intriguing} \cite{goodfellow2015explaining}. Adversarial attacks are carefully designed small perturbations to clean data which significantly change the predictions of target models. The lack of robustness w.r.t adversarial attacks of DNNs brings out security concerns.

In this paper, we focus on defending DNNs from adversarial attacks for image classifications. Many algorithms have been proposed to achieve this purpose. Roughly, those algorithms fall into three categories:
\begin{itemize}
\item data preprocessing, such as JPEG compression \cite{das2017keeping} and image denoise \cite{xu2018feature}.
\item adding stochastic components into DNNs to hide gradient information \cite{athalye2018obfuscated}.
\item adversarial training \cite{madry2018towards}.
\end{itemize}
Data preprocessing or stochastic components are usually combined with adversarial training since it is the most successful defense algorithm.

Recent works show that deep neural networks trained on image classification dataset bias towards textures which are the high-frequency components of images \cite{geirhos2019imagenet-trained}. Meanwhile, researchers empirically find that the perturbations generated by adversarial attacks are also high-frequency signals. This means DNNs are mainly fooled by carefully designed textures. Those facts suggest that suppressing high-frequency components of images is helpful to reduce the effects of adversarial attacks and improve the robustness of DNNs. On the other hand, the basic information on clean images will be retained when suppression high-frequency components because it converges on low frequencies. In this paper, we aim to develop a high-frequency suppressing module which is expected to have the following properties:
\begin{enumerate}
\item \textbf{separability}: it should suppress high-frequency components while keep low-frequency ones.
\item \textbf{efficiency}: it should have low computational costs compared with the standard DNNs.
\item \textbf{differentiability}: it should be differentiable which allows to jointly optimize with adversarial training.
\item \textbf{controllability}: it should be easy to control the degree of high-frequency suppression and the degree of how the original images are modified (e.g. $L_2$ distance).

\end{enumerate}

Discrete Fourier transform (DFT) which maps images into frequency domain is a good tool to achieve those goals. Based on (inverse) DFT, we propose a high-frequency suppressing module which has all those properties. We evaluate our method in the IJCAI-2019 Alibaba Adversarial AI Challenge \cite{IJCAI}. Our code is available on \textcolor{red}{{\small \url{https://github.com/zzd1992/Adversarial-Defense-by-Suppressing-High-Frequencies}}}.

\begin{figure*}[htb]
	\centering
	\subfigure[AAAC]{
		\label{fig:cse_a}
		\includegraphics[width=0.4\textwidth]{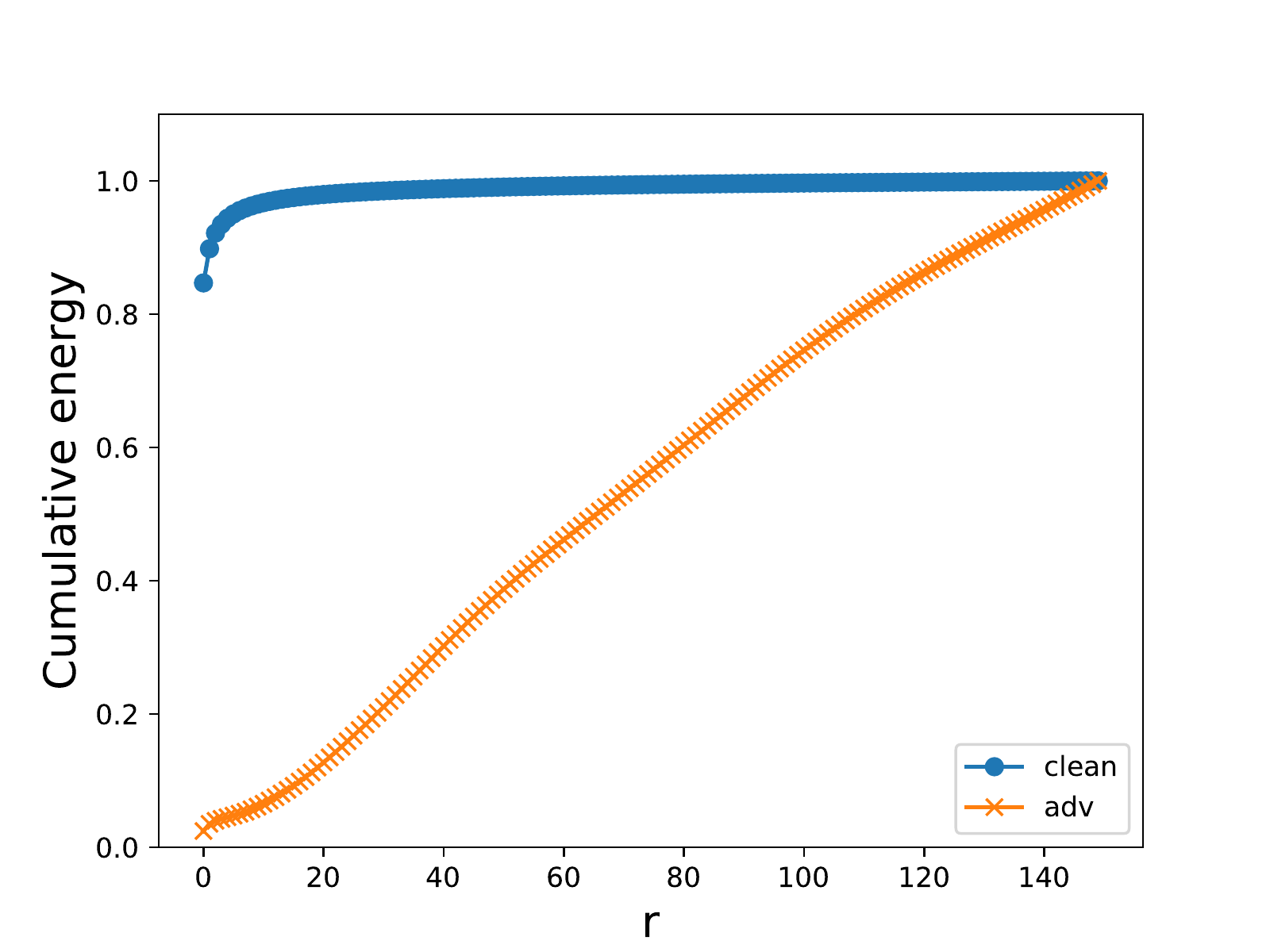}}
	\subfigure[CIFAR-10]{
		\label{fig:cse_b}
		\includegraphics[width=0.4\textwidth]{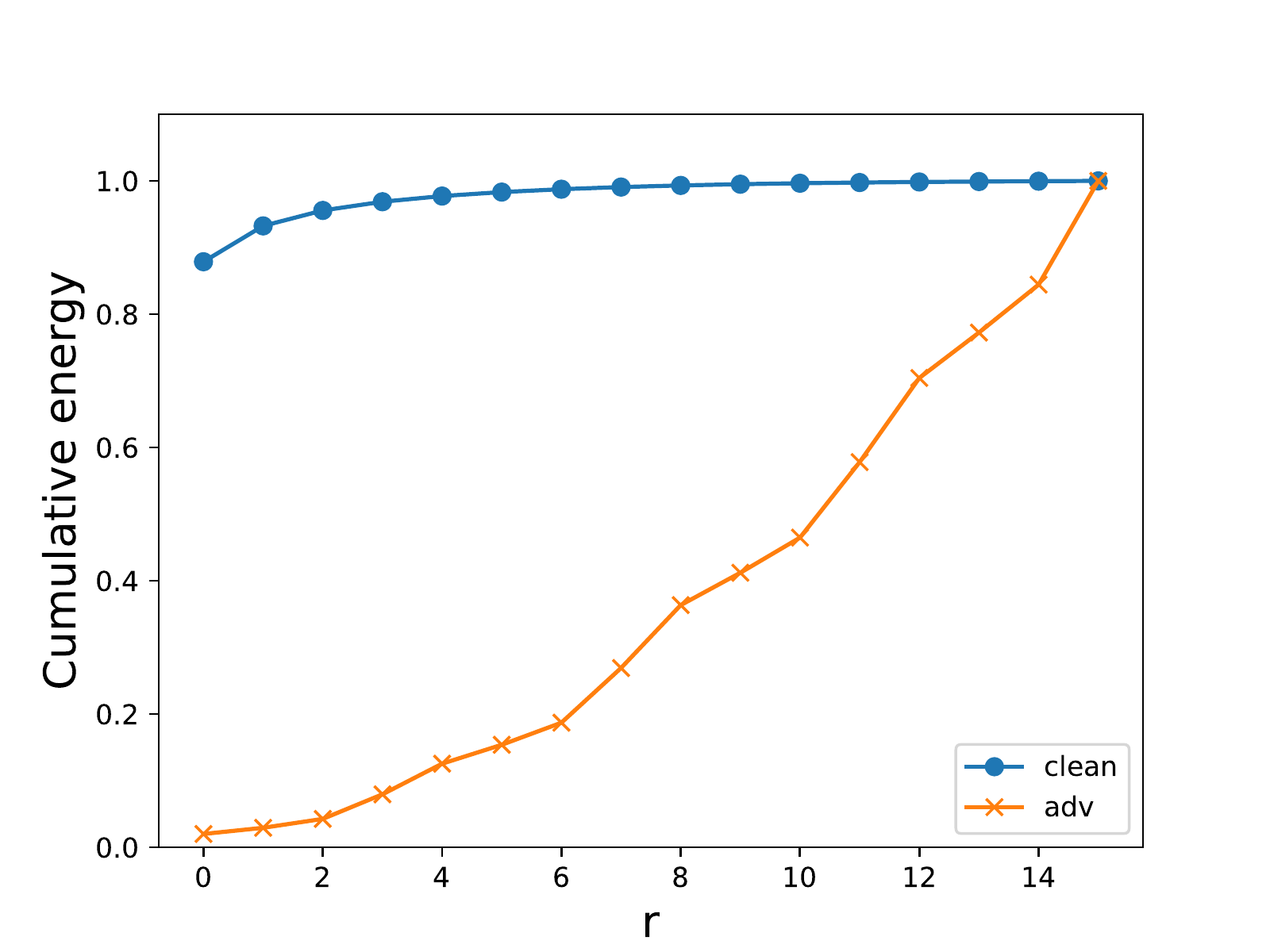}}
	\caption{Cumulative spectrum energy for $5,000$ images of AAAC in (a) and for test images  of CIFAR-10 in (b). Blue line for clean images and orange line for adversarial perturbations.}
\end{figure*}

\section{Method}

\subsection{High-frequency suppression}
As mentioned earlier, suppressing high-frequency components is helpful to reduce the effects of adversarial attacks and improve the robustness of DNNs. Given an input image, we transform it into frequency domain via DFT. Then we reduce the high-frequency components in frequency domain. Finally, we transform the modified frequency image back to time domain.

Formally, denote $\mathbf{x} \in \mathcal{R}^{M\times N}$ as the input image and $\mathbf{\hat{x}} \in \mathcal{C}^{M\times N}$ as its frequency representation.
\begin{equation}
\mathbf{\hat{x}}_{u, v} = \sum_{a=0}^{M-1}\sum_{b=0}^{N-1} \mathbf{x}_{a, b}e^{-j2\pi \left( \frac{u}{M}a+ \frac{v}{N}b\right)}
\end{equation}
To suppress the high-frequency components, we modify $\mathbf{\hat{x}}$ as follows:
\begin{equation} 
\mathbf{\hat{x}} \leftarrow \mathcal{M}  \odot \mathbf{\hat{x}} 
\end{equation}  
where $\mathcal{M} \in \mathcal{R}^{M\times N}$ and $\odot$ is element-wise multiplication. $\mathcal{M}$ controls how different frequency is scaled. Intuitively, $\mathcal{M}$ should close to $0$ for high-frequency components and close to $1$ for low-frequency ones. In this paper, we set $\mathcal{M}$ to a box window with fixed radius $r$. That is
\begin{equation}
\mathcal{M}_{u, v} = \left\{
\begin{array}{lc}
1, \qquad  & 0<=|u|, |v|<=r\\
0, \qquad  & else
\end{array}
\right.
\end{equation}
To simplify the notation, we set $\mathcal{M}_{-u, \cdot} = \mathcal{M}_{M-u, \cdot}$ and $\mathcal{M}_{\cdot, -v} = \mathcal{M}_{\cdot, N-v}$. The overall function of our high-frequency suppression module is
\begin{equation}
\mathbf{x} \leftarrow \mathcal{F}^{-1}\left( \mathcal{M}\odot \mathcal{F}(\mathbf{x})\right)
\end{equation}
where $\mathcal{F}$ means DFT. An image is processed by this module and a standard DNN in order.

Now we analyze the properties of our proposed module.

\textbf{separability}: because $\mathcal{M}$ is a box window, high-frequency components are completely removed and low-frequency ones are perfectly reserved.

\textbf{efficiency}: the computational costs are dominated by DFT. For an $M \times N$ (we suppose $M>=N$) image, the time complexity of DFT is $\mathcal{O}(MN\log_2 M)$. In practice, DFT of a color image is faster than a convolutional layer. Thus the costs of our proposed module are cheap compared with DNNs.

\textbf{differentiability}: DFT can be expressed in matrix form:
\begin{equation}
\mathcal{F}(\mathbf{x}) = \mathbf{F_MxF_N}
\end{equation}
where $\mathbf{F}_M \in \mathcal{C}^{M\times M}$ is the so-called Fourier transform matrix. Clearly, DFT is differentiable. Instead of an image pre-processing method, this property makes it possible to integrate our module into DNNs and optimize with adversarial training jointly. 

\textbf{controllability}: denote $\mathbf{x}_o$ as the output of the proposed module. Based on Parseval theory, we have
\begin{equation}
\lVert \mathbf{x} - \mathbf{x}_o\rVert_2^2= \lVert \mathbf{\hat{x}} - \mathcal{M} \odot \mathbf{\hat{x}}\rVert_2^2
\end{equation} 
Thus the degree of high-frequency suppression and the $L_2$ norm between the original image and the modified image are easily controlled by varying $r$ of the box window. For nature images, spectral energy is converged on low-frequency regions. Thus $\lVert \mathbf{x} - \mathbf{x}_o\rVert_2$ is small enough even when most of the frequency components are suppressed ($r$ is small). 

\subsection{Adversarial training}
The idea of adversarial training is optimizing DNNs w.r.t both clean samples and adversarial samples. 
\begin{equation}
\min_{\mathbf{w}} \left\{ 
\mathcal{L}(f_{\mathbf{w}}(\mathbf{x}), y) + \beta \max_{\lVert \mathbf{\delta} \rVert<\epsilon} \mathcal{L}(f_{\mathbf{w}}(\mathbf{x+\delta}), y)
\right\}
\end{equation}
where $f$ maps an image into classification probability, $\mathbf{w}$ is the parameters of $f$ and $\mathcal{L}$ is the cross-entropy loss. $\mathbf{\delta}$ is obtained by (iteratively) projected gradient descent (PGD). $\beta$ controls the tradeoff between clean samples and adversarial samples.

Recently, \cite{zhang2019theoretically} propose a novel adversarial training method called TRADES. TRADES is formalized as follows:
\begin{equation}
\min_{\mathbf{w}} \left\{ 
\mathcal{L}(f_{\mathbf{w}}(\mathbf{x}), y) + \beta \max_{\lVert \mathbf{\delta} \rVert<\epsilon} \mathcal{L}(f_{\mathbf{w}}(\mathbf{x}), f_{\mathbf{w}}(\mathbf{x+\delta}))
\right\}
\end{equation}
Instead of minimizing the difference between $f_{\mathbf{w}}(\mathbf{x+\delta})$ and the true label, TRADES minimizes the difference between $f_{\mathbf{w}}(\mathbf{x+\delta})$ and $f_{\mathbf{w}}(\mathbf{x})$ which encourages the output to be smooth. In this paper, we use TRADES as the adversarial training method because it has a better tradeoff between robustness and accuracy. Refer \cite{zhang2019theoretically} for more information.

\begin{table*}[htb]
\centering
\Large
\begin{tabular}{cccr}
\hline
High-frequency suppression  & Adversarial training & Model ensemble & Score\\
\hline
$\times$ & $\times$ & $\times$ & 2.0350 \\
$\times$ & $\surd$ & $\times$ & 9.9880 \\
$\surd$ & $\times$ & $\times$ & 14.9736 \\
$\surd$ & $\surd$ & $\times$ & 19.0510 \\
$\surd$ & $\surd$ & $\surd$ & 19.7531 \\
\hline
\end{tabular}
\caption{Ablation study for three strategies and their combinations. }
\label{tab:plain}
\end{table*}

\begin{figure}
	\centering
	\includegraphics[width=0.4\textwidth]{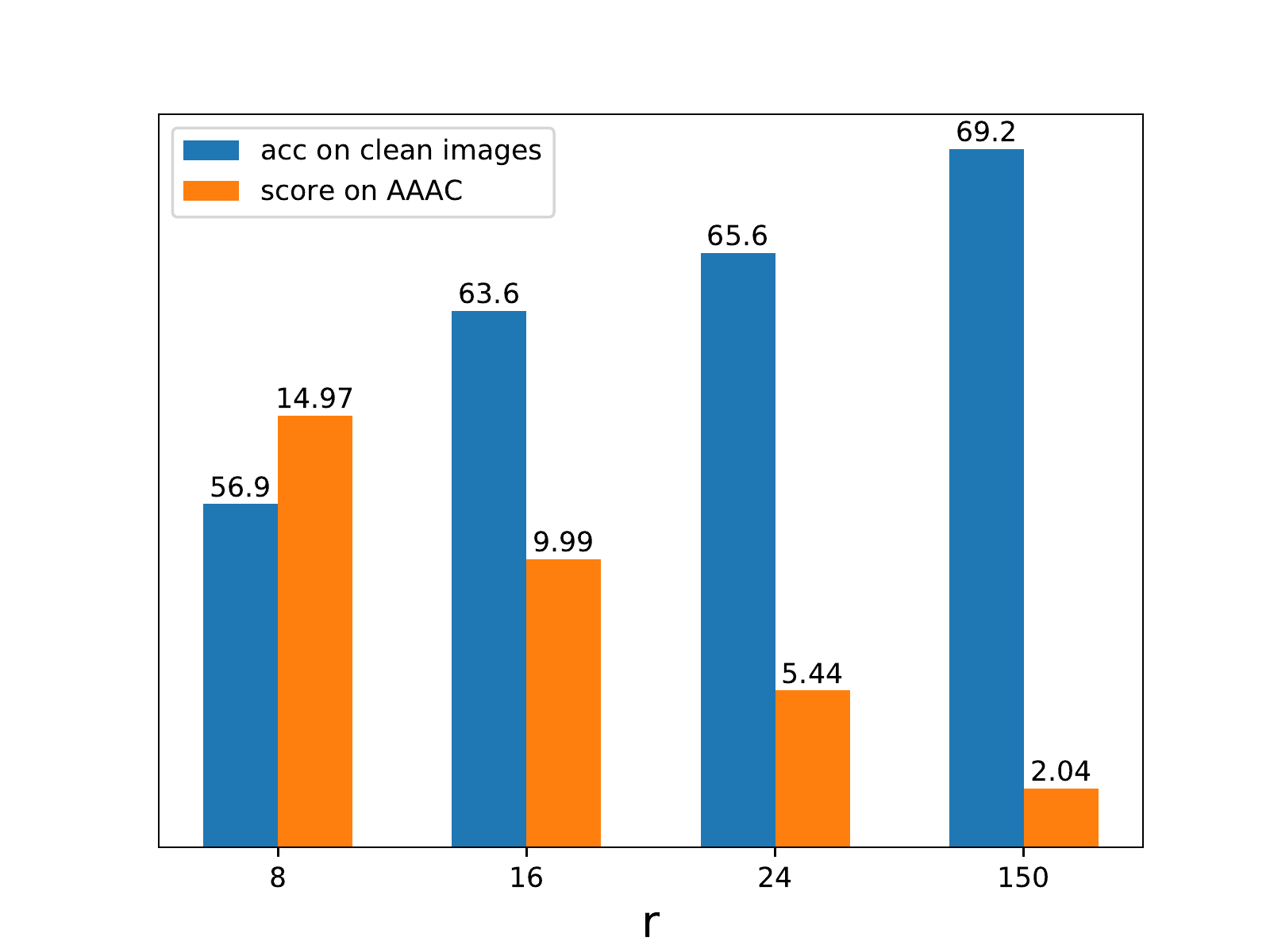}
	\caption{We show the trade-off between robustness and accuracy with high-frequency suppression modules whose $r$ are different. Robustness is measured by the score of AAAC.}
\label{fig:bar}
\end{figure}
\section{Experiments}
We first analyze the statistics of clean images and adversarial images in frequency domain. Then we evaluate the proposed method in the IJCAI-2019 Alibaba Adversarial AI Challenge (AAAC).

In AAAC, models are evaluated by image classification task for electric business. There are about $11,000$ color images from $110$ classes for training. There are $550$ images for test. Given an image, the score of a defense model is calculated as follows:
\begin{equation}
score = \left\{
\begin{array}{cl}
0, \qquad & P_y \neq y \\
mean(\mathbf{\lVert\delta}\rVert_2), \qquad & P_y = y
\end{array}
\right.
\end{equation}
where $P_y$ is the predicted label. The final score is averaged over all images and all black-box attack models. Note that before computing the score, images are resized to $299\times 299$.

We use ResNet-18 as the DNN architecture for all experiments. Our method is implemented with PyTorch.

\subsection{Statistics in frequency domain}
We analyze the statistics of clean samples and adversarial samples in frequency domain. Specifically, we study the distributions of cumulative spectrum energy (CSE) w.r.t frequency. Given a 2D signal $\mathbf{\hat{x}} \in \mathcal{C}^{M\times N}$ in frequecy domain, we define CSE as follows:
\begin{equation}
CSE(r) = \sum_{i=-r}^{r}\sum_{j=-r}^{r} \mathbf{\hat{x}}_{i, j}^* \mathbf{\hat{x}}_{i, j}
\end{equation}
where $r<=\frac{min(M, N)}{2}$. We randomly select $5,000$ images from AAAC. We calculate the CSE score of each image and average all scores. We also calculate the averaged CSE score for the corresponding adversarial perturbations which are generated by iteratively PGD. The results are shown in Fig.~\ref{fig:cse_a}. As we can see, CSE for clean images converges on low-frequency regions while CSE for adversarial perturbations is nearly uniform. Thus, when we suppress the high-frequency components, the effects of adversarial attacks will be significantly reduced while most of the information on clean images will be retained. This is the main motivation of our work.

We calculate CSE score for CIFAR-10, as shown in Fig.~\ref{fig:cse_b}. The distribution is similar to AAAC's. 

\subsection{AAAC results}
As analyzed earlier, when we remove the high-frequency components, the model will be more robust w.r.t adversarial attacks while the accuracy on clean images will be decreased. We evaluate this phenomenon with different $r$ \emph{without} adversarial training. The accuracy is obtained on $5,000$ validation clean images and the robustness is measured by the score of AAAC. We show the results in Fig.~\ref{fig:bar}. As $r$ decreased, the robustness w.r.t adversarial attacks is substantially increased.

Then we do ablation study for three strategies and their combinations: 1) the proposed high-frequency suppression module; 2) adversarial training via TRADES; 3) ensembles of models with different $r$. As we can see in Tab.~\ref{tab:plain}, our proposed module is even better than adversarial training in this challenge and those two methods are complementary to each other. The best score is obtained by ensembling models with different $r$ each of which is trained together with the proposed module and adversarial training. We secure the 5th place in this challenge (the score for the 1st solution is $20.13$).

\section{Conclusions and discussions}
Motived by the difference of frequency spectrum distributions between clean images and adversarial perturbations, we have proposed a high-frequency suppression module to improve the robustness of DNNs. This module is efficient, differentiable and easy to control. We have evaluated our method in AAAC.

We list several directions or questions which are worth to be further explored:
\begin{itemize}
\item Is it helpful to change the radius of box window dynamically?
\item Is it helpful to suppress the high-frequency components of intermediate convolutional features?
\item We evaluate our method for image classification. Does this method work for other tasks or other kinds of data, such as speech recognition?
\end{itemize}

\bibliographystyle{named}
\bibliography{ijcai19}

\end{document}